\RequirePackage{fix-cm}
\documentclass{article}
\PassOptionsToPackage{numbers,sort&compress}{natbib}
\usepackage[preprint]{format_2026}
\usepackage[utf8]{inputenc}
\usepackage[T1]{fontenc}
\DeclareUnicodeCharacter{03C0}{\ensuremath{\pi}}
\DeclareUnicodeCharacter{00B1}{\ensuremath{\pm}}
\usepackage{amsmath,amssymb,amsfonts}
\usepackage{graphicx,booktabs,array,multirow}
\usepackage{microtype,xcolor,url}
\usepackage{placeins,float,needspace}
\usepackage{etoolbox}
\newcommand{\papertablefont}{\fontfamily{ptm}\fontsize{9bp}{10.5bp}\selectfont}
\usepackage[pdfversion=1.7]{hyperref}
\hypersetup{
  colorlinks=true,linkcolor=blue,citecolor=blue,urlcolor=blue,
  pdftitle={LoopVLA: cross-subtask event memory for looped task execution},
  pdfauthor={Shoujing Zhu, Zhenyang Liu, Fungmiu Wong, Jiafeng Wang, Bo Yue, Guiliang Liu, Simo Wu, Xiangyang Xue, Taiping Zeng},
  pdfsubject={Preprint},
  pdfkeywords={vision-language-action, robot manipulation, event memory, looped task execution},
  pdflang={en-US}
}

\title{LoopVLA: cross-subtask event memory\\for looped task execution}

\author{%
\begin{minipage}{\dimexpr\textwidth-2\tabcolsep\relax}\centering
Shoujing Zhu\textsuperscript{1,*,$\ddagger$} \quad
Zhenyang Liu\textsuperscript{1,3,*} \quad
Fungmiu Wong\textsuperscript{1,*} \quad
Jiafeng Wang\textsuperscript{1} \quad
Bo Yue\textsuperscript{2} \\
Guiliang Liu\textsuperscript{2,4} \quad
Simo Wu\textsuperscript{1} \quad
Xiangyang Xue\textsuperscript{1,$\dagger$} \quad
Taiping Zeng\textsuperscript{1,$\dagger$}
\\[6pt]
{\normalfont
\textsuperscript{1}\,Fudan University \quad
\textsuperscript{2}\,School of Data Science, The Chinese University of Hong Kong, Shenzhen \quad
\textsuperscript{3}\,Shanghai Innovation Institute \quad
\textsuperscript{4}\,Shenzhen Loop Area Institute
\\[4pt]
{\small\texttt{\{zhusj25, liuzy24, fmwang23, jfwang25\}@m.fudan.edu.cn, boyue@link.cuhk.edu.cn,}} \\
{\small\texttt{liuguiliang@cuhk.edu.cn, wusimo19911125@gmail.com}}\\
{\small\texttt{ \{xyxue, zengtaiping\}@fudan.edu.cn}}
\\[4pt]
{\footnotesize
\textsuperscript{*}\,Equal contribution \quad
\textsuperscript{$\ddagger$}\,Project Lead \quad
\textsuperscript{$\dagger$}\,Corresponding author}%
}
\end{minipage}
}
\date{}

\begin{document}
\raggedbottom
\maketitle
\AtBeginEnvironment{tabular}{\papertablefont}
\begin{abstract}
Repetitive tasks are common in real-world manipulation, yet current vision-language-action (VLA) policies remain unreliable for repeated actions. We present LoopVLA, a cross-subtask event-memory interface for looped task execution. Its central idea is to compress a just-completed subtask into memory for the next subtask's action expert, making completed execution available for subsequent repetition and stopping decisions. Learnable event latents query the closed segment's tokens, while action features query history and joint history--event context; the two action-side readouts are residually fused and injected directly into the action head. On a memory benchmark, LoopVLA achieves $80.22\%$ success on the repetition task group with online subgoal prediction, bringing evaluation closer to realistic deployment while improving overall success over strong memory baselines.
\end{abstract}

\section{Introduction}

Vision-Language-Action (VLA) policies such as RT-2~\citep{brohan2023rt2}, OpenVLA~\citep{kim2024openvla}, $\pi_0$/$\pi_{0.5}$~\citep{black2024pi0,physicalintelligence2025pi05}, and GR00T~N1.x~\citep{nvidia2025gr00t,nvidiagear2025gr00tn16} generate manipulation actions from visual observations, robot state, and task instructions. Active perception~\citep{liu2026activevla} and visual grounding with spatial reasoning~\citep{liu2025reasongrounder,liu2025neural} improve scene understanding, yet the next action also depends on completed execution: visually similar states may call for another repetition or for stopping.

This ambiguity is central to \emph{history-dependent manipulation}. A packaging station must place the requested number of items, an assembly cell must repeat an operation across parts, and an inspection fixture must stop an object after a prescribed number of movements. In such looped tasks, successful execution requires both performing each subtask and deciding when its repetitions are complete. The resulting \emph{cross-subtask dependency} calls for memory that transfers completed execution to the next action decision while retaining access to earlier progress.

Recent memory-augmented VLAs address temporal context through multi-observation summaries and visual traces~\citep{koo2025hamlet,zheng2024tracevla,fang2025sam2act,liu2025trivla}, semantic or episodic memory and stage retrieval~\citep{torne2026mem,sridhar2025memer,li2025mapvla}, and recurrent or query-based summaries~\citep{shi2025memoryvla,li2026rememvla}. These approaches differ in what they store, when they update or access it, and where it conditions control. Subgoal control~\citep{seqvla2025,streamvla2026,han2025robocerebra} and keyframe-based history~\citep{bpp2026,kcvla2026} further motivate event-level temporal abstraction. We study a specific cross-subtask interface: at a subgoal transition, learnable event latents compress the most recently closed execution segment into memory that the next subtask's action expert reads alongside historical tokens.

We identify a key design principle: \textbf{subtask transitions are natural moments to organise recently completed execution for the next stage.} Sampled perceptual history retains earlier context, while an event summary represents the most recently closed segment. The next subtask's action expert reads these complementary memories to inform continuing, repeating, or stopping.

Based on this principle, we present \textbf{LoopVLA}, a cross-subtask event-memory interface for looped task execution. In the compression stage, \emph{learnable event latents} $Q^e$ serve as queries~\citep{li2023blip2}, and the closed segment's event tokens provide keys and values. Attention aggregation produces content-dependent event memory $m_{k-1}$. In the action-reading stage, action hidden states separately query historical tokens and the joint history--event context containing $m_{k-1}$. The two readouts undergo residual fusion before the fused context is injected through MemoryRMS into the $\pi_{0.5}$ action expert. The task instruction $\ell$ remains a policy input; subsequent-subtask action supervision and auxiliary subgoal alignment train the memory to preserve information useful for execution.

\begin{figure}[t]
\centering
\includegraphics[width=\linewidth]{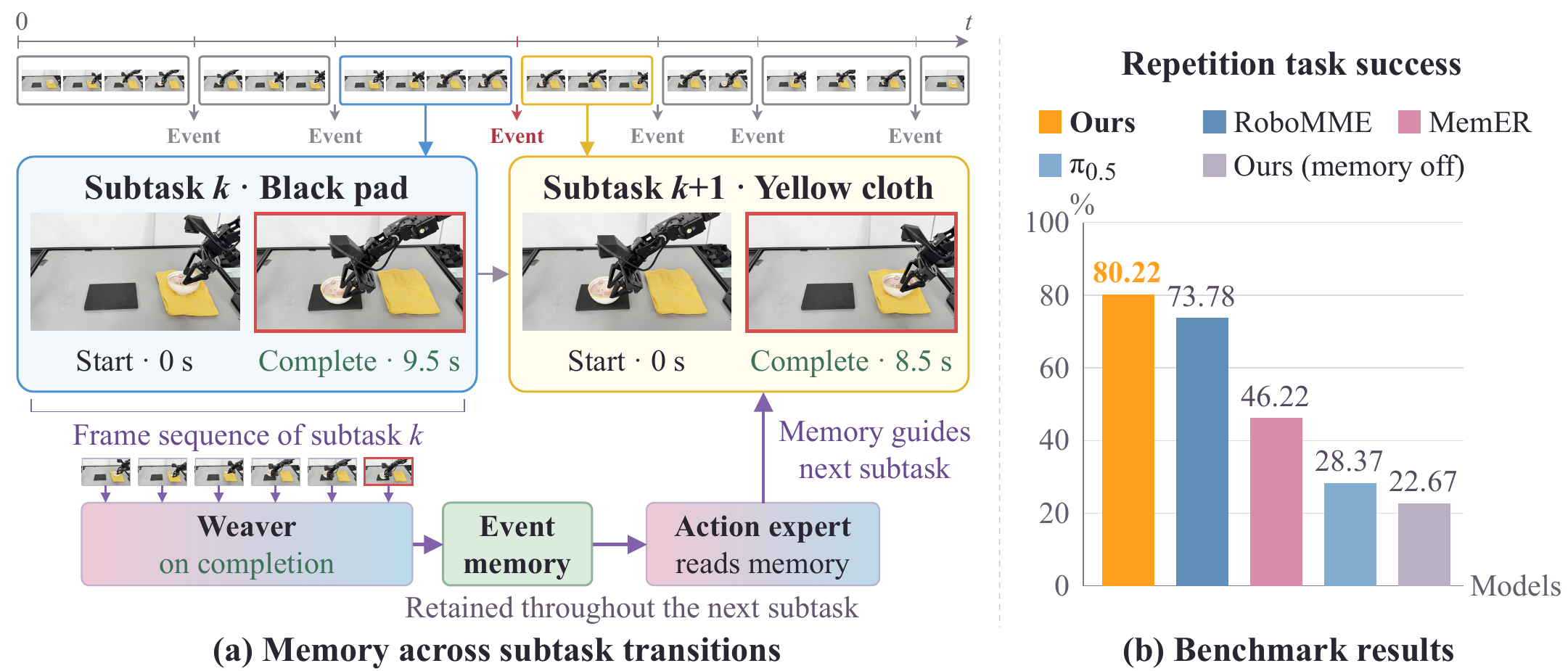}
\caption{\textbf{Memory connects completed subtasks to subsequent actions.} (a) A timeline of a robotic task. Blue segments are compressed into event memory to guide actions in the yellow segment; events trigger memory compression. (b) Results on repeated tasks.}
\label{fig:trigger-storyboard}
\end{figure}

Figure~\ref{fig:trigger-storyboard} illustrates this memory hand-off and compares success on repeated tasks (in panel~(b), Ours' results use online Qwen-VL subgoals to approximate a real-world setting). RoboMME (FrameSamp+Modul)~\citep{dai2026robomme} is the baseline shown in panel~(b). Table~\ref{tab:main_results} compares the best-performance configuration with strong memory baselines. Across all 16 tasks, LoopVLA achieves $47.17\%$ success with Qwen-VL subgoals and $47.71\%$ with Oracle subgoals, compared with $44.51\%$ for RoboMME (best), the strongest published baseline in this comparison. On the repetition task group---\textsc{PickXtimes}, \textsc{SwingXtimes}, and \textsc{StopCube}---the corresponding rates are $80.22\%$, $79.78\%$, and $73.78\%$; on \textsc{StopCube}, they are $49.33\%$, $48.67\%$, and $42.00\%$. Reference-configuration analyses further examine window organisation, history coverage, repetition-count and difficulty strata, and paired episode outcomes. Both subgoal protocols are formal execution evaluations: Oracle uses simulator annotations, while Qwen-VL predicts subgoals online to bring evaluation closer to realistic deployment.

Our contributions are:
\begin{enumerate}\setlength\itemsep{1pt}
\item We identify subtask completion as a natural, task-aligned point for memory compression, highlighting the \emph{when} dimension of cross-subtask hand-off. Each self-contained summary directly conditions the next subtask's action expert, specifying \emph{where} memory is consumed.
\item We propose LoopVLA, whose compact Weaver summarises completed subtasks into event memory, defining \emph{what} to pass forward. Action-driven reading and residual fusion combine this memory with history for direct injection into the action expert.
\item On a memory benchmark, LoopVLA uses online subgoal prediction to approximate a real-world setting and achieves $80.22\%$ aggregate success on the repetition task group, versus $22.67\%$ for the memory-off configuration. This demonstrates effective repeated execution with predicted event triggers.
\end{enumerate}

\section{Related work}
\label{sec:related}

\subsection{Memory-augmented VLA policies}

Memory-augmented VLA policies differ along three axes: \emph{what} is stored, \emph{where} it is consumed, and \emph{when} it is updated or accessed. Prior work stores visual histories or overlays~\citep{koo2025hamlet,zheng2024tracevla,fang2025sam2act,liu2025trivla}, semantic traces or retrieved episodes~\citep{torne2026mem,sridhar2025memer,li2025mapvla}, and recurrent/query summaries~\citep{shi2025memoryvla,li2026rememvla}; these memories are consumed through prompt/input augmentation~\citep{li2025mapvla,jang2025contextvla,fan2025longvla}, planning or hierarchical control~\citep{sridhar2025memer,zhang2024hirt}. LoopVLA targets the `what, where and when' by compressing the most recently closed execution segment into event memory supplied directly to the next sub-task's action expert.

Along the \emph{when} axis, observation-level updates~\citep{koo2025hamlet}, retrieval of demonstration-time stages~\citep{li2025mapvla}, and rollout-time sub-goal events organise memory access differently. Completion-aware execution~\citep{seqvla2025,streamvla2026}, keyframe-based history~\citep{bpp2026,kcvla2026}, hierarchical sub-goal control~\citep{han2025robocerebra}, and skill segmentation in imitation learning~\citep{kipf2019compile,zhu2022buds,wan2024lotus} provide related temporal abstractions. In LoopVLA, an observed sub-goal transition closes the current execution segment; its event memory is then passed to the next sub-task's action expert.

\subsection{Query-bottleneck memory modules}

Learnable query bottlenecks compress variable-length context into fixed-size representations; BLIP-2's Q-Former~\citep{li2023blip2} uses learnable queries to bridge frozen vision and language modules. Related representation work includes hierarchical robot control~\citep{zhang2024hirt}, cross-segment recurrent memory~\citep{bulatov2022rmt}, and latent sub-segments in imitation learning~\citep{kipf2019compile}. Our trigger--weaver factorisation is inspired by MemGen~\citep{zhang2025memgen}, whose trigger decides when to invoke a latent-memory weaver during LLM reasoning. LoopVLA adapts this division of roles to robot control: sub-goal transitions close execution segments, whose event tokens provide keys and values, while learnable event latents serve as queries. Attention pooling produces content-dependent event memory for the next sub-task's action expert; Section~\ref{sec:weaver} gives the formulation.

\subsection{Task-structure priors and repetition counting}

Task-structure methods track progress through reward-machine states~\citep{toroicarte2018rewardmachines} or LTL formula progression in instruction-conditioned policies~\citep{vaezipoor2021ltl2action}. LoopVLA instead represents the closed segment with learned event memory shaped by subsequent-action prediction and sub-goal alignment. Perception-side work studies repetition counting in video~\citep{dwibedi2020repnet,hu2022transrac,sinha2024escounts} and object counting in vision-language models~\citep{paiss2023countclip,vlmcountbench2025}. LoopVLA addresses the complementary control question of how past execution should condition the next decision to repeat, proceed, or stop. Its learnable event latents produce a representation for subsequent control, without assuming that this representation encodes an exact, decodable repetition count.

\section{LoopVLA architecture}

\subsection{Overview}

\begin{figure}[t]
\centering
\includegraphics[width=\linewidth]{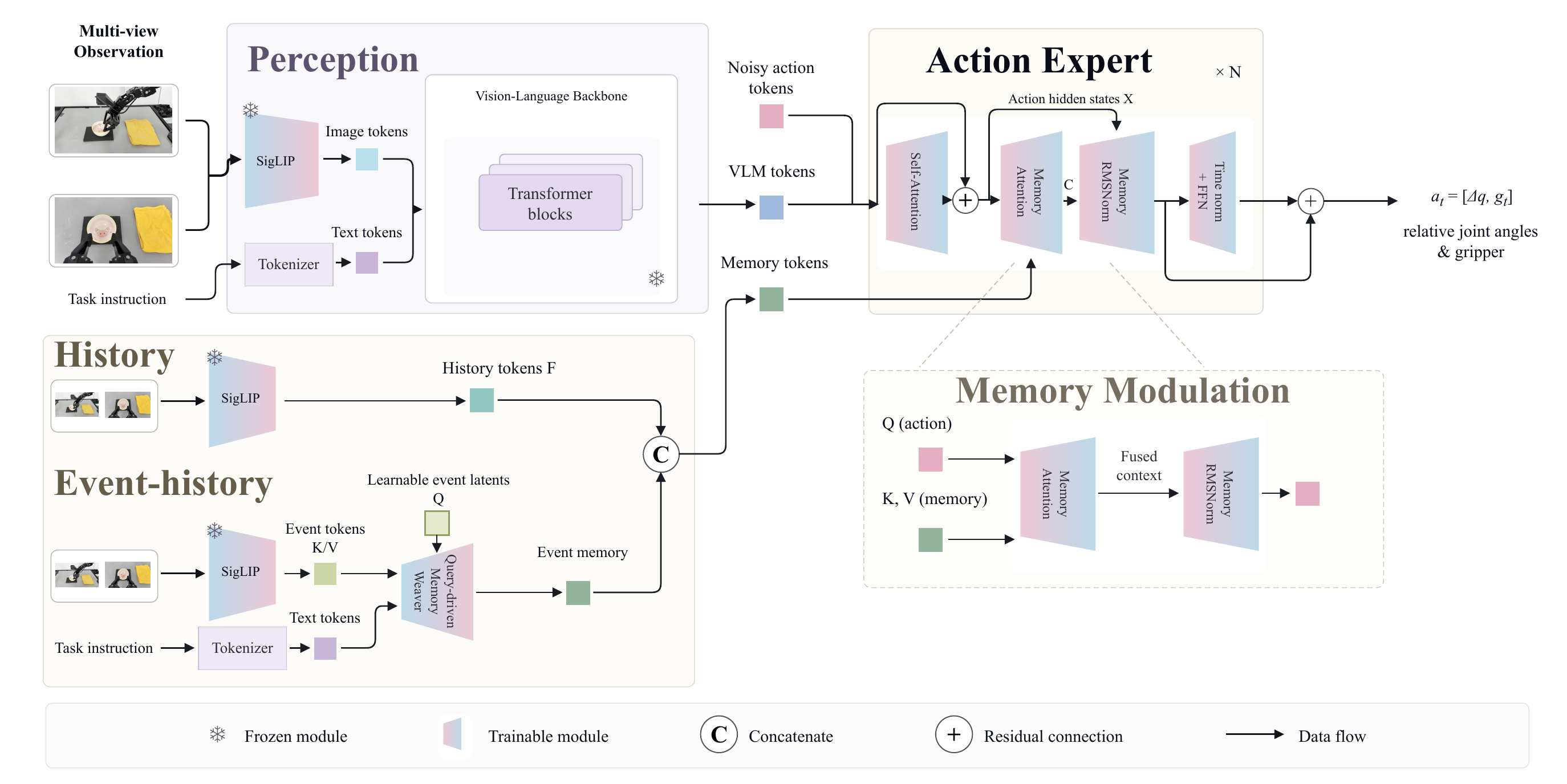}
\caption{\textbf{History and event memory are consumed by the action expert.} Learnable event latents query segment event tokens (K/V) to form event memory. Action hidden states read history and joint history--event context; residual fusion and output projection supply the context injected through MemoryRMS. Each square denotes a token set; snowflakes denote frozen modules.}
\label{fig:framework}
\end{figure}

LoopVLA targets ordered sub-tasks $\mathcal{E}=(\tau_0,\ldots,\tau_{K-1})$ whose actions depend on earlier execution. The VLA predicts an action chunk~\citep{zhao2023act,liu2025spatial} from observation $o_t$, state $s_t$, and instruction $\ell$, augmented with perceptual history $F_t=\operatorname{History}(o_{0:t})$ and the preceding segment's summary. Learnable event latents $Q^e$ query segment event tokens $Z_{k-1}$ to produce the compressed output $m_{k-1}$. The joint memory is $M_t=[F_t\Vert m_{k-1}]$. At each policy query,
\begin{equation}
\begin{aligned}
m_{k-1} &\leftarrow
\begin{cases}
\operatorname{Weaver}(Z_{k-1};Q^e,\ell), & \text{on an observed event transition},\\
m_{k-1}, & \text{otherwise},
\end{cases}\\
\hat a_t&\leftarrow\operatorname{ActionExpert}(o_t,s_t,\ell;\underbrace{M_t}_{\makebox[0pt]{$\scriptstyle\operatorname{History}(o_{0:t})\Vert m_{k-1}$}}).
\end{aligned}
\label{eq:overview}
\end{equation}
Action hidden states separately read $F_t$ and joint memory $M_t$; residual fusion supplies the context for MemoryRMS (Section~\ref{sec:consumption} and Figure~\ref{fig:framework}). The visual branches in Figure~\ref{fig:framework} use SigLIP~\citep{zhai2023siglip}.

The event memory is initially empty; closing a segment updates $m_{k-1}$, which remains available throughout the next segment $\tau_k$. History $F_t$ advances with execution, and current action hidden states initiate each read. The available summary can therefore yield different readouts at successive actions.

\subsection{Perceptual history}
\label{sec:history}

We uniformly sample up to $H$ observations across the executed history through the current timestep; encoding and projection produce $F_t$. The main memory setting is 32 history observations + 32 event latents ($H=M=32$). Appendix~\ref{appx:hparam} describes event inputs, encoding, and padding.

\subsection{Sub-goal-event trigger}
\label{sec:trigger}

A change in the sub-goal label supplied to the policy closes the preceding segment and updates event memory. Let $g_q$ denote this label at query $q$. Queries occur every 16 executed actions, so an event is an observed change in $g_q$. This defines the segment for memory hand-off, complementing observation-level summaries~\citep{koo2025hamlet} and demonstration-stage retrieval~\citep{li2025mapvla}. Appendix~\ref{appx:hparam} summarises event inputs and updates.

To approximate realistic deployment, we use \emph{Qwen-VL} to predict sub-goals online and supply labels to this event interface. We also evaluate an \emph{Oracle} protocol using simulator-annotated sub-goals as a controlled benchmark setting. Section~\ref{sec:experiments} reports task outcomes under both protocols. Figure~\ref{fig:trigger-storyboard} illustrates how a completed subtask supplies event memory to the next subtask's action expert.

Event boundaries provide a temporal abstraction related to option termination~\citep{sutton1999options,bacon2017optioncritic}. Each summary describes the most recently closed segment and remains available until the next observed transition; earlier execution remains accessible through $F_t$.

\subsection{Query-driven memory weaver}
\label{sec:weaver}

For the preceding segment $\tau_{k-1}$, the Weaver encodes up to $n=130$ observations into event tokens $Z_{k-1}\in\mathbb{R}^{L\times d_h}$ using frozen visual, temporal-position, and proprioceptive features, projection, and layer normalisation~\citep{ba2016layernorm}. Learnable event latents $Q^e\in\mathbb{R}^{M\times d_h}$ provide queries; $Z_{k-1}$ supplies keys and values. Attention pooling~\citep{vaswani2017attention} normalises over segment tokens to compress variable-length input into $M$ fixed slots. With episode-instruction conditioning implicit in $Z_{k-1}$ and $Q^e$,
\begin{equation}
\begin{aligned}
K^e&=V^e=Z_{k-1},\\
A^e&=\operatorname{softmax}\!\left(\frac{Q^e(K^e)^\top}{\sqrt{d_h}}+B^e\right),\\
m_{k-1}&=P_{\mathrm{out}}(A^eV^e+Q^e).
\end{aligned}
\label{eq:weaver}
\end{equation}
Here $B^e$ masks segment padding, and $P_{\mathrm{out}}$ maps to memory width $d$ with output normalisation. The trainable slots $Q^e$ have an internal query residual $+Q^e$; only their content-dependent output $m_{k-1}\in\mathbb{R}^{M\times d}$ enters the action-side K/V context alongside $F_t$. This output describes the latest closed segment; $F_t$ supplies earlier execution. Appendix~\ref{appx:hparam} details inputs, conditioning, and empty-segment handling.

\subsection{Action-side memory consumption}
\label{sec:consumption}

We instantiate the interface in the Gemma action expert~\citep{gemmateam2024gemma,gemmateam2024gemma2} of $\pi_{0.5}$~\citep{physicalintelligence2025pi05}. Let $X_j$ be the action features after block $j$'s self-attention residual. Define $\operatorname{Read}_j(X,S)$ as the attention readout before its output projection, with queries from $X$ and keys and values from $S$. The action features read history and joint history--event context:
\begin{subequations}
\label{eq:memxattn}
\begin{align}
R_j^{\mathrm{hist}}&=\operatorname{Read}_j(X_j,F_t),\label{eq:readhist}\\
R_j^{\mathrm{joint}}&=\operatorname{Read}_j(X_j,[F_t\Vert m_{k-1}]).\label{eq:readjoint}
\end{align}
$R_j^{\mathrm{hist}}$ and $R_j^{\mathrm{joint}}$ are action-dependent reads, each normalised within its valid context. The reads use RMSNorm~\citep{zhang2019rmsnorm} and RoPE~\citep{su2024roformer}.

Residual fusion combines the two readouts before the output map produces $C_j$:
\begin{equation}
C_j=P_j^O\!\left(\operatorname{ResFuse}_j(R_j^{\mathrm{hist}},R_j^{\mathrm{joint}})\right).
\label{eq:resfuse}
\end{equation}
\end{subequations}
MemoryRMS takes action features $X_j$ and context $C_j$, which generates their scale and shift:
\begin{subequations}
\label{eq:memrms}
\begin{align}
[\gamma_j,\beta_j]&=C_jW_j^\Delta+b_j^\Delta,\label{eq:scaleshift}\\
X_j'&=\frac{X_j}{\sqrt{\operatorname{mean}(X_j^2)+\varepsilon}}
\odot(1+\gamma_j)+\beta_j.\label{eq:modulation}
\end{align}
\end{subequations}
The Weaver's query residual and the action-side residual fusion are separate operations. MemoryRMS precedes the existing timestep-conditioned pre-FFN norm and FFN, preserving the original time-conditioning pathway.

\subsection{Training strategy and objective}
\label{sec:staged}

The vision encoder and non-adapter language parameters of PaliGemma~\citep{beyer2024paligemma} remain frozen. Language LoRA adapters~\citep{hu2022lora}, the action expert, history projection, Weaver, memory attention/modulation, and alignment projection are trained. Training pairs each preceding segment's summary with subsequent-segment actions, restricting history to observations available at each action.

We retain the base policy's flow-matching action objective~\citep{lipman2022flowmatching}, conditioned on the joint memory $M_t$:
\begin{equation}
\mathcal{L}_{\text{action}}
=\mathbb{E}_{\rho,a,\epsilon}\,
\bigl\|v_\theta(a^{(\rho)},\rho\mid o_t,s_t,\ell,M_t)-(\epsilon-a)\bigr\|_{2,\mathrm{valid}}^2.
\label{eq:actionloss}
\end{equation}
The norm averages squared error only over valid action-horizon positions, excluding padding beyond the episode end.

An auxiliary cosine-alignment objective $\mathcal{L}_{\mathrm{align}}$ aligns the event summary with the upcoming sub-goal. The total objective is
\begin{equation}
\mathcal{L}=\mathcal{L}_{\mathrm{action}}+\lambda_{\mathrm{align}}\mathcal{L}_{\mathrm{align}}.
\label{eq:training}
\end{equation}
Subsequent-stage action supervision trains the learnable event latents by backpropagating through action reads, residual fusion, and MemoryRMS. Semantic alignment supplies a training target for the event representation; predicted sub-goals supply labels for online event triggering. Appendix~\ref{appx:hparam} lists the key training settings.
\FloatBarrier

\section{Experiments}
\label{sec:experiments}

\FloatBarrier
\subsection{Experimental setup}
\paragraph{Tasks and reporting.}
We evaluate all 16 RoboMME tasks~\citep{dai2026robomme}, with 50 benchmark episodes per task and evaluation seed. We report equal-weight task means for all 16 tasks and the \emph{repetition task group}: \textsc{PickXtimes}, \textsc{SwingXtimes}, and \textsc{StopCube}. Requested repetition count $N$ and Easy/Medium/Hard difficulty are separate strata. Means and sample SD summarize evaluation seeds 7, 21, and 42 of one fixed checkpoint. Each protocol contains 2,400 episodes; group SD uses seed-level group means.

\paragraph{Observation and event protocols.}
Both protocols use the same checkpoint of our method (Ours) and memory configuration, with nominal subgoal queries every 16 executed actions. For deployment-oriented evaluation, Qwen3-VL-4B-Instruct~\citep{bai2025qwen3vl} supplies online subgoals using the simple-subgoal adapter; post-hold label changes trigger memory updates. Oracle supplies simulator-annotated subgoals on the same query grid as a second formal evaluation protocol. Both use seeds 7, 21, and 42. History and window configuration comparisons use Oracle and common seed 7. Timing analysis uses all 2,400 Qwen-VL trajectories, with simulator labels on visited states as temporal references. Configuration and training settings appear in Appendices~\ref{appx:configurations} and~\ref{appx:training}.

\FloatBarrier
\subsection{Performance against prior work}
\label{sec:agg}\label{sec:repeat-results}
We compare with $\pi_{0.5}$, DP-Unet~\citep{chi2023diffusionpolicy}, SAM2Act~\citep{fang2025sam2act}, MemER~\citep{sridhar2025memer}, and RoboMME (FrameSamp+Modul), the strongest non-SG RoboMME variant on repetition tasks. We label this variant \emph{RoboMME (best)} in the tables. Table~\ref{tab:main_results} shows that LoopVLA with Qwen-VL reaches $80.22\%$ on repetition tasks, compared with $73.78\%$ for RoboMME (best). It also improves overall success and performs best on each of the three repetition tasks, supporting the use of completed-subtask memory for repeated execution. Table~\ref{tab:task-comparison} in the appendix compares all seven methods on every task. Public baselines retain their published aggregation, so these comparisons are descriptive.
\begin{table}[!htbp]
\centering\footnotesize
\setlength{\tabcolsep}{3pt}
\caption{Task success (\%). The repetition task group averages PickXtimes, SwingXtimes, and StopCube. Public baselines follow RoboMME aggregation~\citep{dai2026robomme}; Ours reports mean ± sample SD over evaluation seeds 7, 21, and 42. Column-best means, including ties, are red.}
\label{tab:main_results}
\begin{tabular}{cccccc}
\toprule
Method & All 16 & Repetition & PickX & SwingX & StopCube \\
\midrule
π\textsubscript{0.5} & 17.93 & 28.37 & 42.89 & 35.56 & 6.67 \\
DP-Unet & 7.02 & 4.00 & 5.33 & 6.67 & 0.00 \\
SAM2Act & 21.37 & 33.78 & 76.00 & 25.33 & 0.00 \\
MemER & 42.38 & 46.22 & 79.33 & 59.33 & 0.00 \\
RoboMME (best) & 44.51 & 73.78 & 87.33 & 92.00 & 42.00 \\
\textbf{Ours (Qwen-VL)} & 47.17 ± 1.42 & \textcolor[HTML]{C73535}{80.22 ± 2.04} & \textcolor[HTML]{C73535}{94.00 ± 2.00} & \textcolor[HTML]{C73535}{97.33 ± 2.31} & \textcolor[HTML]{C73535}{49.33 ± 6.43} \\
\textbf{Ours (Oracle)} & \textcolor[HTML]{C73535}{47.71 ± 1.37} & 79.78 ± 1.39 & 93.33 ± 1.15 & \textcolor[HTML]{C73535}{97.33 ± 1.15} & 48.67 ± 3.06 \\
\bottomrule
\end{tabular}
\end{table}

\FloatBarrier
\subsection{History coverage for repeated execution}
\label{sec:history-coverage}
Table~\ref{tab:history-results} compares $H\in\{10,20,32,40\}$ with $M=32$, Oracle, and seed 7. Success increases from $39.88\%$ at $H=10$ to $49.25\%$ at $H=32$, while success on repetition tasks increases from $60.00\%$ to $81.33\%$. Increasing to $H=40$ gives $46.75\%$ and $74.00\%$, respectively. Thus $H=32$ has the highest success among the compared configurations; additional coverage does not continue to improve execution. Among the latent counts we tried, 32 event latents yielded the best performance; additional latents did not improve execution. Appendix~\ref{appx:configurations} retains all four points in a separate history figure and records their metadata.
\begin{table}[!htbp]
\centering\footnotesize
\setlength{\tabcolsep}{3pt}
\caption{History-coverage configurations under Oracle, seed 7. All rows use 32 event latents; column-best success rates are red.}
\label{tab:history-results}
\begin{tabular}{cccc}
\toprule
History observations & Event latents & All 16 (\%) & Repetition (\%) \\
\midrule
10 & 32 & 39.88 & 60.00 \\
20 & 32 & 41.75 & 66.67 \\
32 & 32 & \textcolor[HTML]{C73535}{49.25} & \textcolor[HTML]{C73535}{81.33} \\
40 & 32 & 46.75 & 74.00 \\
\bottomrule
\end{tabular}
\end{table}

\begin{figure}[!htbp]
\centering
\includegraphics[width=\linewidth]{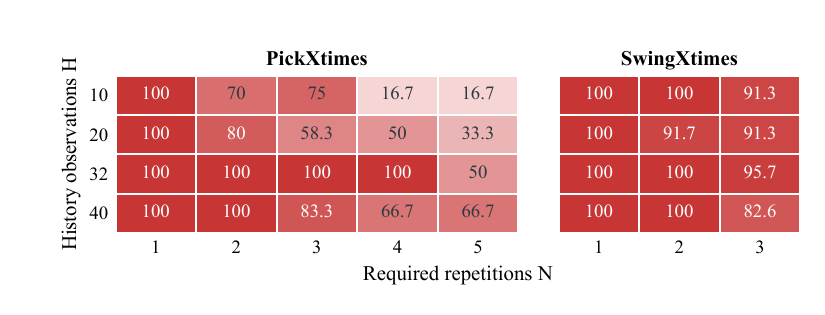}
\caption{\textbf{History coverage across repetition demands.} Success (\%) under Oracle, seed 7, $M=32$. Cells share a white-to-red 0--100\% scale and display their rates. These are configuration comparisons of history observations.}
\label{fig:history-demands}
\end{figure}

Figure~\ref{fig:history-demands} resolves these aggregate rates by required repetition count. All four configurations complete every single-repeat episode in both tasks. They separate as the demand grows: at four PickXtimes repetitions, success counts are 1/6, 3/6, 6/6, and 4/6 for increasing $H$. At five repetitions, $H=40$ reaches 4/6 versus 3/6 at $H=32$. The aggregate preference for $H=32$ therefore coexists with a remaining high-count limitation; the full grid retains both patterns.

\FloatBarrier
\Needspace{8\baselineskip}
\subsection{Event-aligned windows improve task completion}
\label{sec:window}
\begin{figure}[!htbp]
\centering
\includegraphics[width=\linewidth]{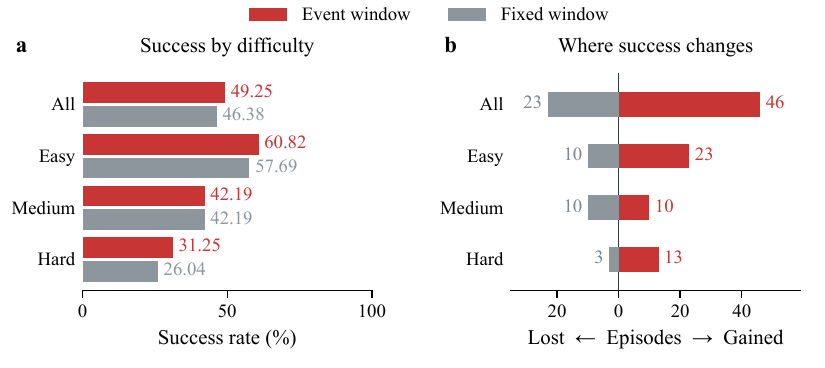}
\caption{\textbf{Event-aligned windows expand successful execution.} (a) Absolute success rates under Oracle, seed 7; $H=M=32$. (b) Newly successful (red) and no-longer-successful (gray) benchmark keys relative to the fixed window.}
\label{fig:window-difficulty}
\end{figure}

We compare the most recently closed event segment with a fixed recent window. Both use 32 history observations + 32 event latents and an event-input cap of 130 observations. Under Oracle and seed 7, event-aligned windows improve success from $46.38\%$ to $49.25\%$, including an increase from $26.04\%$ to $31.25\%$ on Hard tasks. Figure~\ref{fig:window-difficulty} shows both rates and changes in episode outcomes.

Across 800 matched benchmark keys, the event configuration gains 46 successes and loses 23, with 10 net gains on Hard tasks. This supports using a complete event segment for subsequent actions, while the configuration differences prevent an isolated causal claim. Appendix~\ref{appx:window-transitions} describes the pairing.

\FloatBarrier
\subsection{Online subgoals support repeated execution}
\label{sec:n-strat}
Both protocols achieve $100\%$ on \textsc{PickXtimes} through $N=4$; at $N=5$, Qwen-VL and Oracle reach $50.00\pm16.67\%$ and $44.44\pm9.62\%$. At $N=3$ on \textsc{SwingXtimes}, they reach $97.10\pm2.51\%$ and $94.20\pm2.51\%$.

\FloatBarrier
\Needspace{8\baselineskip}
\subsection{Strong task performance with online subgoals}
\label{sec:predictor-quality}
\begin{figure}[H]
\centering
\includegraphics[width=\linewidth]{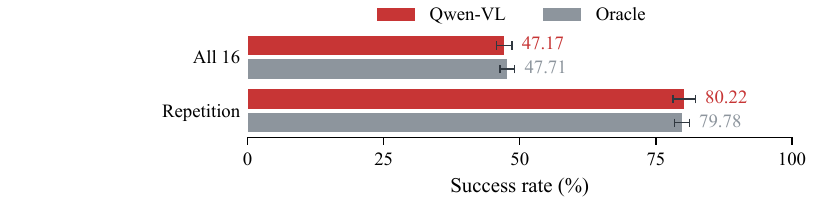}
\caption{\textbf{Strong task performance with online subgoals.} Mean $\pm$ sample SD over evaluation seeds 7, 21, and 42; 2,400 episodes per protocol, including 450 episodes from repetition tasks.}
\label{fig:predictor-quality}
\end{figure}

Qwen-VL predicts subgoals online to approximate realistic deployment. Figure~\ref{fig:predictor-quality} compares its execution performance with Oracle. Across 2,400 matched rollouts, Qwen-minus-Oracle is $-0.54$ percentage points overall (95\% CI $[-1.67,0.50]$) and $+0.44$ on repetition tasks ($[-1.78,2.44]$). Each of the three group tasks differs by at most $0.67$ points; Appendix~\ref{appx:full-tables} reports all tasks. These intervals describe protocol differences for this checkpoint and evaluation set.

\paragraph{Continuous execution between memory updates.}
Across all 2,400 Qwen-VL trajectories, ordered event F1 is $88.86\%$; matched boundaries have median offset $+10$ actions. Perceptual history continues updating between boundaries, while the last closed-segment memory remains available for action-conditioned reads until replacement. Our model is robust under these conditions and executes actions in a real-world setting. Matching has no timing cutoff; Appendix~\ref{appx:predictor-quality} explains event F1 and the estimation procedure.

\FloatBarrier
\subsection{Real-world repeated execution}
\label{sec:real-world}
We train LoopVLA with 200 real-robot demonstrations per task and deploy it on two NVIDIA RTX 5090 GPUs, using Qwen-VL subgoals for online memory updates. We evaluate LoopVLA and $\pi_{0.5}$ on 20 trials per task: \emph{Drawer Push--Pull} (three pull--push cycles), \emph{Bowl Transfer} (four round trips), and \emph{Battery Transfer} (three round trips followed by placement in a cup). Each task requires tracking completed cycles to stop or switch actions at the requested count.

LoopVLA succeeds in 39 of 60 real-robot trials, compared with 21 for $\pi_{0.5}$ (Table~\ref{tab:real-world-results}). These results show that cross-subtask event memory also supports repeated execution in real-world settings.
\begin{table}[!htbp]
\centering
\caption{Real-robot task success (successes/trials), with 20 trials per task and method.}
\label{tab:real-world-results}
\begin{tabular}{ccccc}
\toprule
\textbf{Method} & \shortstack{\textbf{Drawer}\\\textbf{Push\textendash Pull}} & \shortstack{\textbf{Bowl}\\\textbf{Transfer}} & \shortstack{\textbf{Battery}\\\textbf{Transfer}} & \textbf{Total} \\
\midrule
π\textsubscript{0.5} & 5/20 & 7/20 & 9/20 & 21/60 \\
\textbf{Ours} & \textcolor[HTML]{C73535}{\textbf{11/20}} & \textcolor[HTML]{C73535}{\textbf{13/20}} & \textcolor[HTML]{C73535}{\textbf{15/20}} & \textcolor[HTML]{C73535}{\textbf{39/60}} \\
\bottomrule
\end{tabular}
\end{table}

\FloatBarrier

\section{Conclusion}

We presented \textbf{LoopVLA}, a cross-subtask event-memory interface that organises completed execution at subgoal transitions for the next subtask's action expert. Learnable event latents summarise the most recently closed segment; action-dependent reads of perceptual history and joint history--event context are fused and injected through MemoryRMS to inform when to continue, repeat, or stop.

With online Qwen-VL subgoals, LoopVLA achieves $47.17\%$ overall success across 16 simulation tasks and $80.22\%$ on repetition tasks, versus $22.67\%$ with memory off. Real-robot results show that this memory mechanism also supports repeated execution in real-world settings, with 39 successes in 60 trials for LoopVLA and 21 for $\pi_{0.5}$.

\label{page:main-end}

\FloatBarrier
\begingroup
\renewcommand{\bibfont}{\small}
\bibliographystyle{plainnat}
\bibliography{references}
\endgroup

\clearpage
\appendix
\section{Implementation Details}
\label{appx:hparam}\label{appx:training}

Table~\ref{tab:hparams} lists the settings needed to interpret the experiments.

\begin{table}[htbp]
\centering\small
\caption{Key settings for LoopVLA.}
\label{tab:hparams}
\begin{tabular}{>{\centering\arraybackslash}p{0.30\linewidth}>{\centering\arraybackslash}p{0.64\linewidth}}
\toprule
Component & Setting \\
\midrule
Policy backbone & π\textsubscript{0.5} with a trainable memory interface \\
Memory setting & 32 history observations + 32 event latents \\
Event input & Up to 130 observations from the latest closed segment \\
Event query & Every 16 executed actions \\
Action prediction & Chunks of 20 actions; episode-end padding is masked \\
Training & 3 epochs; AdamW with warm-up and cosine decay \\
Training objectives & Flow matching and next-subgoal cosine alignment \\
\bottomrule
\end{tabular}
\end{table}

\paragraph{Event inputs and updates.}
Visual, temporal-position, and proprioceptive features are projected and normalised. The instruction conditions the event features and learnable queries. Padding is masked; an empty segment produces invalid, zero-valued memory. Offline subgoal labels are carried between query points, and each transition pairs the closed segment with subsequent actions. History uses only observations available at each action. Online, Qwen-VL predicts the subgoal labels used to trigger event-memory updates.

\section{Complete Numerical Results}
\label{appx:statistics}\label{appx:full-tables}\label{appx:agg-table}

\paragraph{Aggregation and coverage.}
Each task has 50 benchmark episodes per policy seed. All-task success is the equal-weight mean over 16 tasks; the repetition task group averages PickXtimes, SwingXtimes, and StopCube. Both protocols use seeds 7, 21, and 42 of the same fixed checkpoint and report their mean and sample SD ($\mathrm{ddof}=1$). Group SD is calculated from each seed's group mean.

Table~\ref{tab:task-comparison} compares the seven main-table methods on every task. All LoopVLA results use the same fixed best-performance checkpoint.

\begin{table}[!htbp]
\centering\footnotesize
\setlength{\tabcolsep}{2.5pt}
\caption{Task success (\%). Ours reports means over evaluation seeds 7, 21, and 42; public baselines follow RoboMME aggregation~\citep{dai2026robomme}. The three repetition tasks are listed first. Row-best means, including ties, are red.}
\label{tab:task-comparison}
\begin{tabular}{cccccccc}
\toprule
Task & π\textsubscript{0.5} & DP-Unet & SAM2Act & MemER & \shortstack{RoboMME\\(best)} & \shortstack{\textbf{Ours}\\\textbf{(Qwen-VL)}} & \shortstack{\textbf{Ours}\\\textbf{(Oracle)}} \\
\midrule
\textbf{PickXtimes} & 42.89 & 5.33 & 76.00 & 79.33 & 87.33 & \textcolor[HTML]{C73535}{94.00} & 93.33 \\
\textbf{SwingXtimes} & 35.56 & 6.67 & 25.33 & 59.33 & 92.00 & \textcolor[HTML]{C73535}{97.33} & \textcolor[HTML]{C73535}{97.33} \\
\textbf{StopCube} & 6.67 & 0.00 & 0.00 & 0.00 & 42.00 & \textcolor[HTML]{C73535}{49.33} & 48.67 \\
\midrule
BinFill & 30.00 & 30.67 & 40.00 & \textcolor[HTML]{C73535}{56.67} & 39.56 & 40.67 & 44.67 \\
VideoUnmask & 20.44 & 14.00 & 27.33 & \textcolor[HTML]{C73535}{81.33} & 32.67 & 43.33 & 42.67 \\
ButtonUnmask & 22.22 & 4.00 & 32.00 & \textcolor[HTML]{C73535}{72.00} & 25.11 & 26.67 & 22.00 \\
VideoUnmaskSwap & 18.67 & 4.00 & 18.00 & \textcolor[HTML]{C73535}{38.00} & 24.44 & 26.67 & 29.33 \\
ButtonUnmaskSwap & 6.67 & 0.00 & \textcolor[HTML]{C73535}{26.67} & 21.33 & 18.22 & 11.33 & 19.33 \\
PickHighlight & 11.33 & 0.00 & 17.33 & \textcolor[HTML]{C73535}{70.67} & 22.89 & 24.00 & 23.33 \\
VideoRepick & 0.44 & 0.00 & 5.33 & 25.33 & 30.44 & 30.00 & \textcolor[HTML]{C73535}{30.67} \\
VideoPlaceButton & 31.11 & 22.00 & 24.67 & 30.00 & \textcolor[HTML]{C73535}{60.00} & 55.33 & 55.33 \\
VideoPlaceOrder & 25.78 & 4.33 & 20.00 & 26.00 & 32.00 & \textcolor[HTML]{C73535}{39.33} & 38.67 \\
MoveCube & 26.00 & 8.00 & 29.33 & 82.67 & 77.78 & 82.67 & \textcolor[HTML]{C73535}{84.00} \\
InsertPeg & 1.56 & 0.00 & 0.00 & 6.67 & \textcolor[HTML]{C73535}{7.56} & 6.67 & 6.67 \\
PatternLock & 2.89 & 3.33 & 0.00 & 16.67 & 53.56 & 61.33 & \textcolor[HTML]{C73535}{62.67} \\
RouteStick & 4.67 & 10.00 & 0.00 & 12.00 & \textcolor[HTML]{C73535}{66.67} & 66.00 & 64.67 \\
\midrule
\textbf{All 16} & 17.93 & 7.02 & 21.37 & 42.38 & 44.51 & 47.17 & \textcolor[HTML]{C73535}{47.71} \\
\textbf{Repetition} & 28.37 & 4.00 & 33.78 & 46.22 & 73.78 & \textcolor[HTML]{C73535}{80.22} & 79.78 \\
\bottomrule
\end{tabular}
\end{table}

\paragraph{Memory-off results.}
\label{appx:memory-off}
The memory-off configuration succeeds in 19/50 \textsc{PickXtimes}, 14/50 \textsc{SwingXtimes}, and 1/50 \textsc{StopCube} episodes, giving $22.67\%$ across this group. The main Qwen-VL protocol succeeds in 361/450 episodes ($80.22\%$).

\paragraph{Configuration metadata and comparability.}
\label{appx:configurations}\label{appx:window}\label{appx:diff}
Table~\ref{tab:docx_history_results} records the memory settings used in the history and window comparisons. All use the same event-input cap, Oracle subgoals, and evaluation seed 7.

\begin{table}[htbp]
\centering\footnotesize
\setlength{\tabcolsep}{3pt}
\caption{Memory settings for the history and window comparisons.}
\label{tab:docx_history_results}
\begin{tabular}{ccc}
\toprule
Configuration & History observations & Event latents \\
\midrule
Event history & 10 & 32 \\
Event history & 20 & 32 \\
Main event history & 32 & 32 \\
Event history & 40 & 32 \\
Fixed recent window & 32 & 32 \\
\bottomrule
\end{tabular}
\end{table}

\begin{figure}[!htbp]
\centering
\includegraphics[width=\linewidth]{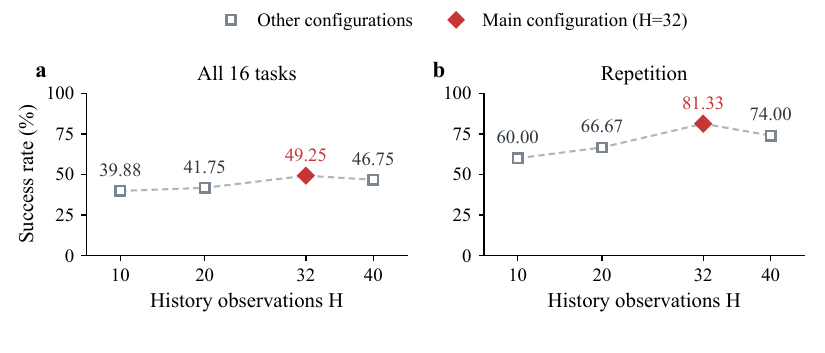}
\caption{\textbf{History coverage across all compared configurations.} Oracle, seed 7, $M=32$: 800 episodes per point, including 150 episodes from repetition tasks. The $H=32$ configuration achieves the highest measured success. All four budgets are shown on 0--100\% axes; points are single-seed rates. Table~\ref{tab:docx_history_results} documents configuration differences.}
\label{fig:history-coverage}
\end{figure}

\paragraph{Window pairing.}
\label{appx:window-transitions}
The window comparison pairs 800 task/episode keys at seed 7. This assumes the same initial states.

\FloatBarrier
\section{Execution with Online Subgoals}
\label{appx:predictor-quality}

\paragraph{Execution with predicted subgoals.}
Table~\ref{tab:deployment-summary} shows that Qwen-VL achieves repetition-task success close to Oracle despite imperfect boundary timing. Perceptual history continues to update, and the previous event memory remains available until replacement. These results support robust execution with online subgoal predictions.

\begin{table}[!htbp]
\centering\footnotesize
\setlength{\tabcolsep}{5pt}
\caption{Execution and event timing with online Qwen-VL subgoals in simulation. Offsets use matched boundaries; positive values indicate later updates.}
\label{tab:deployment-summary}
\begin{tabular}{ccc}
\toprule
Scope & Metric & Result \\
\midrule
All 16 tasks & Ordered event F1 & 88.86\% \\
 & Median boundary offset & +10 actions \\
\midrule
Repetition tasks & Success (Qwen-VL / Oracle) & 80.22\% / 79.78\% \\
\bottomrule
\end{tabular}
\end{table}

\paragraph{What event F1 measures.}
Ordered event F1 is the harmonic mean of precision and recall for subgoal transitions matched by stage label and temporal order. Extra predictions lower precision; missed reference transitions lower recall. The score of $88.86\%$ indicates good recovery of the subgoal sequence. Matching has no timing cutoff, so F1 measures transition recognition; boundary offsets separately measure timing accuracy.

\paragraph{Measurement.}
Timing analysis uses all 2,400 Qwen-VL trajectories. The protocol intervals in Section~\ref{sec:predictor-quality} use 5,000 task-stratified bootstrap resamples of benchmark episode keys, retaining both protocols and all three seeds together. They describe this checkpoint and evaluation set, not variation across training runs.

\FloatBarrier

\Needspace{5\baselineskip}
\section{Execution Examples and Real-world Setup}
\label{appx:weaver-comparison-frames}
\setlength{\intextsep}{6pt plus 2pt minus 2pt}

\begin{figure}[h!]
\centering
\includegraphics[width=0.95\linewidth]{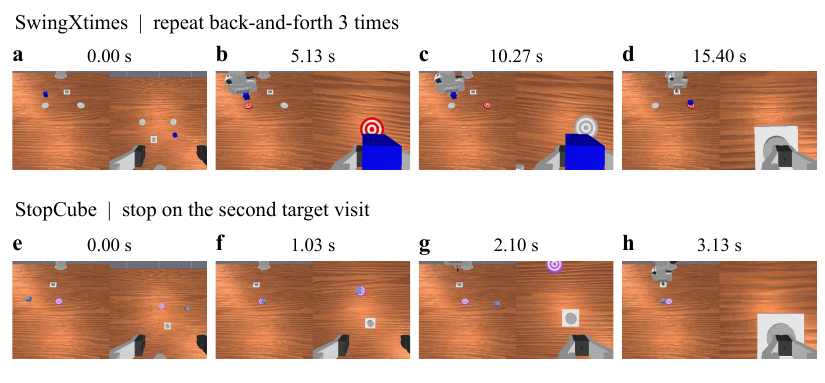}
\caption{\textbf{Archived successful rollouts with predicted sub-goals.} Earlier reference configuration, checkpoint 21284, Qwen-VL, seed 7. Top: SwingXtimes episode 17, three back-and-forth repetitions. Bottom: StopCube episode 3, stopping on the second target visit. Each image retains both camera views. Timestamps are rendered-video seconds; images are uniformly sampled.}
\label{fig:rollout-cases}
\end{figure}

\begin{figure}[!htbp]
\centering
\includegraphics[width=0.95\linewidth]{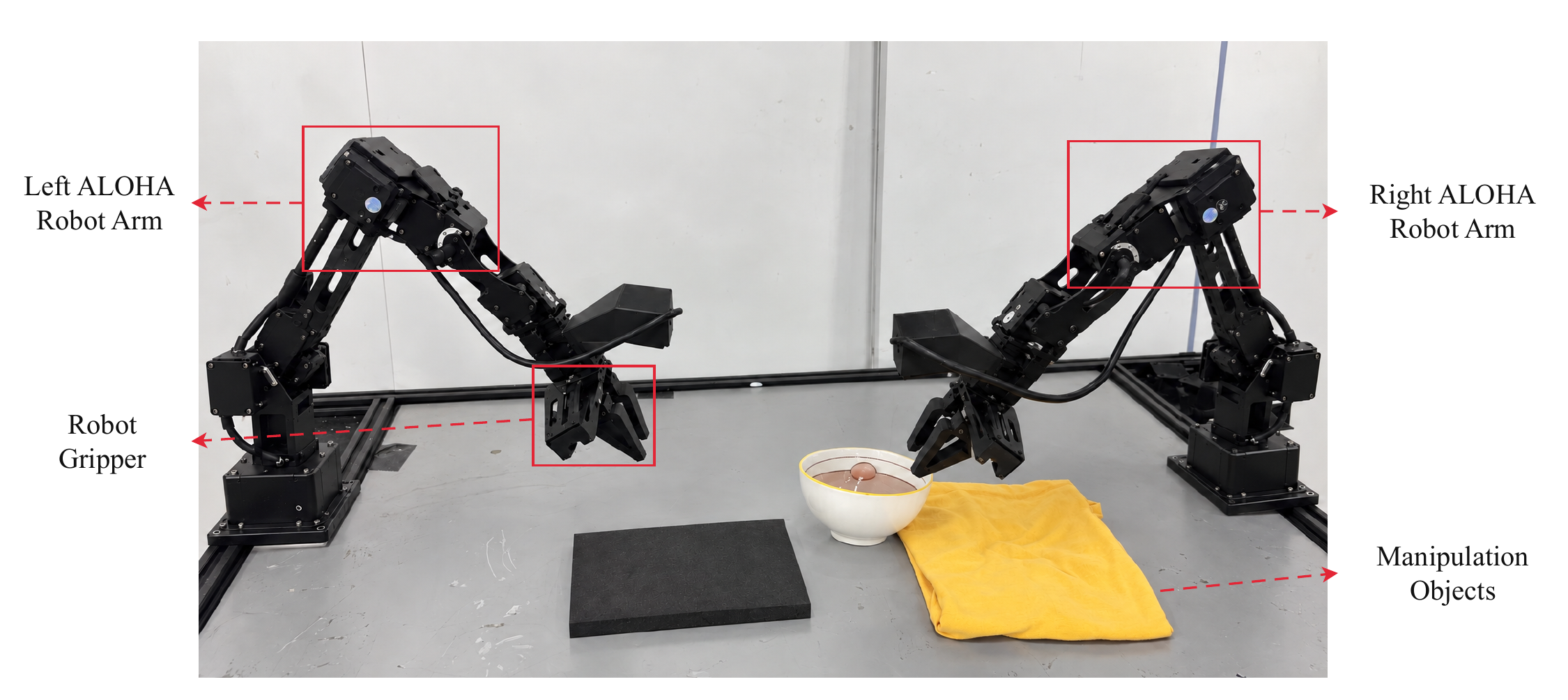}
\caption{\textbf{Real-world experimental setup.} Illustration of the dual-arm ALOHA platform and tabletop workspace used for the repeated manipulation tasks.}
\label{fig:real-world-setup}
\end{figure}

\paragraph{Real-world setup.}
Figure~\ref{fig:real-world-setup} illustrates the dual-arm ALOHA setup. Using online Qwen-VL subgoals, LoopVLA succeeds in 39 of 60 real-robot trials, compared with 21 for π\textsubscript{0.5} (Section~\ref{sec:real-world}). These results show that the memory mechanism also works for repeated execution in a physical setting.

\FloatBarrier

\end{document}